___________________________________________

*Does AlphaGo actually play Go?*
**Concerning the State Space of Artificial Intelligence**

___________________________________________

Holger Lyre
Chair of Theoretical Philosophy &
Center for Behavioral Brain Sciences
University of Magdeburg, Germany
lyre@ovgu.de

**Abstract**

The overarching goal of this paper is to develop a general model of the state space of AI. Given the breathtaking progress in AI research and technologies in recent years, such conceptual work is of substantial theoretical interest. The present AI hype is mainly driven by the triumph of deep learning neural networks. As the distinguishing feature of such networks is the ability to self-learn, self-learning is identified as one important dimension of the AI state space. Another main dimension lies in the possibility to go over from specific to more general types of problems. The third main dimension is provided by semantic grounding. Since this is a philosophically complex and controversial dimension, a larger part of the paper is devoted to it. We take a fresh look at known foundational arguments in the philosophy of mind and cognition that are gaining new relevance in view of the recent AI developments including the blockhead objection, the Turing test, the symbol grounding problem, the Chinese room argument, and general use-theoretic considerations of meaning. Finally, the AI state space, spanned by the main dimensions generalization, grounding and "self-x-ness", possessing self-x properties such as self-learning, is outlined.

# 1. Introduction

There is much to suggest that 15 March 2016 should be retrospectively regarded as a historical date. On this day Lee Sedol, one of the strongest Go players in the world, loses the last game of a tournament lasting several days against the AI system "AlphaGo" of the development company Google DeepMind. AlphaGo defeated the South Korean with 4:1 games. The event attracted worldwide attention and brought back memories of the victory of the IBM computer program "Deep Blue" against the then reigning world chess champion Garri Kasparov some 20 years earlier in 1997. And yet the similarity of both events is rather superficial.

For a long time, especially in the sixties and seventies, chess was considered the testbed and benchmark for the development of machine intelligence. In the nineties, however, not only had this awe gone, but almost all experts had also seen the victory of DeepBlue coming. Even though chess has impressive combinatorics, it was obvious that sooner or later computers would be able to calculate the decision tree more efficiently and further ahead than any human player due to rapidly increasing computing power. DeepBlue owed its success essentially to brute computational power, so that even its developers were very reluctant to voice their opinions on AI's superiority over humans. AlphaGo's victory, on the other hand, came almost out of nowhere. Just a few years earlier, computers were at best capable of playing Go at amateur level. The complexity of Go is vastly greater than that of chess. It is said that Go is to chess as chess is to checkers. In fact, the victory of AlphaGo was a real surprise for large parts of the professional world.

DeepBlue and AlphaGo use radically different architectures. While DeepBlue is a classic, symbolic and rule-based AI system, AlphaGo is based on a deep learning neural network (DL network). The two systems are thus paradigmatic examples of the two longstanding and rivaling schools of symbolism and connectionism, of "Good Old Fashioned Artificial Intelligence" (GOFAI) and neural networks. DL networks belong to the latest development in neural network research (see LeCun etal. 2015). They are called "deep" as they consist of more than just two or three, sometimes even hundreds of layers. The breathtaking successes of DL applications in the last ten years have led to what Sejnowski (2018) calls the "deep learning revolution". What makes these systems special and what in fact distinguishes virtually all neural networks since the perceptron in the late 1950s is their ability to learn, or, even stronger and in view of the recent developments, to *self-learn* by actively exploiting or exploring big training data or by self-interaction with virtual or real environments.

There can be no doubt that the deep learning revolution has led to a new hype in AI over the recent ten years, be it in science, industry, economy or the media. These developments provide a strong motivation to rethink the question of what constrains the evolution of AI understood as the general quest to develop thinking machines or artificial minds. The idea of the paper is that a closer look at the deep learning revolution combined with a fresh look at foundational philosophy of mind questions in the light of the new developments will guide us in our search for the dimensions that span the general state space of AI. Hence, the ultimate goal of the paper is this: to develop and propose a model of the state space of AI.

The paper proceeds as follows. In the second section, the key ingredients of the deep learning revolution that drives the advances and successes of AI over the last ten years will be revisited. The discussion will provide us with a proposal for two important dimensions that span the state space of AI: self-learning and generalization. A further main dimension consists in semantic grounding. The discussion of this philosophically complex and controversial topic will occupy the third and largest section of the paper. We shall take a fresh look at well-known foundational arguments in the philosophy of mind and cognition that are gaining new relevance in view of the recent AI developments such as the blockhead objection, the Turing test, the symbol grounding problem, the Chinese room argument, and general use-theoretic considerations of meaning. In the fourth and final section, a general model of the state space of AI will be proposed and outlined. It shall be argued that the AI state space is spanned by the main dimensions generalization, grounding and self-x-ness, where grounding and self-x-ness decompose into further sub-dimensions (particularly the concept of self-x-ness originates from the movement of organic computing and includes self-learning as a special case).

## 2. The Deep learning revolution

### 2.1 Self-learning AI

AlphaGo and DeepBlue use radically different architectures. In contrast to DeepBlue, AlphaGo is based on a DL network (see LeCun etal. 2015 and Schmidhuber 2015a, 2015b for overviews). The learning-based training of the network proceeded in two steps. In its initial phase, AlphaGo first learned patterns and moves of human Go players from a database of millions of moves. It was then able to give move recommendations similar to those of experts. In a second training phase, the system learned by self-play on the basis of reinforcement learning. This second phase demonstrates the crucial qualitative difference from AlphaGo to DeepBlue: the ability to self-learn.

Classical GOFAI builds on the assumption that intelligence and cognition are nothing but rule-based manipulations of symbols. DeepBlue is a classic AI system par excellence in this regard. To calculate the evaluation function of millions of possible positions based on a given starting position (on average more than 100 million positions per second) DeepBlue relied on the expert knowledge of numerous chess grandmasters implemented in the calculation algorithms. As a non-learning system it was only able to operate within the framework of the given implementation. Such a limitation is typical for a classical GOFAI architecture: DeepBlue was designed for one special purpose, and was therefore unable to perform any other task than playing chess. It is a specialized or „narrow" AI (ANI: artificial narrow intelligence).

The more recent AI development almost reverses the original GOFAI doctrine. It is precisely the self-learn ability that opens up the field of flexible general intelligence. It is, conversely, rather blocked by the GOFAI concept of full rule-based programming. In retrospect, it seems hard to understand how the importance of learning could have been downplayed in the early stages of AI. Paradigmatic for this view is the position of Noam Chomsky (1980), according to which the human language ability is not otherwise understandable than under the assumption of a presupposed, allegedly innate deep grammar, i.e. a deeply anchored rule competence that is universal to humans. Chomsky considered it out of the question that such an ability could have arisen through imitation or reinforcement learning. Terrence Sejnowski comments on this very clearly:

> *"What is innate is not grammar, but the ability to learn language from experience and to absorb the higher-order statistical properties of utterances in a rich cognitive context. What Chomsky could not imagine was that, when coupled with deep learning of the environment and a deeply learned value function honed by a lifetime of experience, a weak learning system like reinforcement learning can indeed give rise to cognitive behaviors, including language. ...it follows logically from Orgel's second rule: evolution is cleverer than you are, and that includes experts like Chomsky."* (Sejnowski 2018, S. 251)

A few more distinctions about the notion of learning are in order. Machine learning (ML) in general is about developing algorithms to extract regularities or patterns from training data.

This leads to learning a function that maps an input to an output. Learning can be either supervised, unsupervised or consist of elements of both. Reinforcement learning is the standard example for an in-between case. Learning algorithms are said to be supervised if both the input and the desired output are given. Backpropagation is probably the most well-known type of a neural network algorithm for supervised learning. The idea is that the difference between the network output and the supervisory teaching signal is used to backpropagate a rule for adjusting the weights of the network. Learning is unsupervised when only the inputs are given. In this case the system has to learn to extract regularities or patterns from the input data in a self-organized manner. Main methods in ML are principle component analysis and cluster analysis. Important types of unsupervised neural networks are self-organizing maps such as Kohonen networks or the Willshaw-Malsburg model. Reinforcement learning is typically considered as a third type of learning besides supervised and unsupervised learning. Here, learning improves on the basis of feedback in terms of reward for good performance (see the standard textbook by Sutton & Barto 2018 for an overview).

A notion of special interest that somewhat crosscuts the above distinctions is the notion of *self-learning*. While it is sometimes used as synonymous to unsupervised learning, self-learning in a more general sense should be understood as the system's ability to exploit or explore the training data by itself, or, as in the case of AlphaGo's self-play, to even generate them. This would also include most cases of reinforcement learning, as the learning reward presupposes that such systems actively interact with their evironments (whether artificial or real). Demis Hassabis, co-founder and CEO of DeepMind, talks in the same way about self-learning.[1] Yann LeCun, one of the pioneers of the DL revolution, recently proposed to distinguish the category of "self-supervised" learning from the weaker unsupervised learning and the even weaker reinforcement learning: "in self-supervised learning, the system learns to predict parts of its input from other parts of its input".[2] Self-supervised learning is thus largely congruent with the general concept of self-learning introduced here (the latter comprising not only self-generated but also self-explored inputs).

---

1  Demis Hassabis: The Power of Self-Learning Systems. Talk at MIT Center for Brains, Minds, and Machines, March 20, 2019. URL: https://cbmm.mit.edu/news-events/events/cbmm-special-seminar-self-learning-systems
2  Yann LeCun, April 30, 2019, on Twitter https://twitter.com/ylecun/status/1123235709802905600 and Facebook https://www.facebook.com/722677142/posts/10155934004262143/

## 2.2 Classical AI, new AI, and the three waves of connectionism

There is an interesting meaning shift in the term "AI" as it is used nowadays in connection with the recent AI hype compared to its use over decades in large parts of the scientific community. At least for the last 30 years of the 20$^{th}$ century, the term AI had mainly been used to indicate classical AI or GOFAI. This doctrine, again, is based on the paradigm of symbolism, i.e. the assumption that cognition is rule-based symbol manipulation. This is mainly a software requirement, the specific hardware remains secondary, if not irrelevant. Connectionism, on the other hand, represents an opposite paradigm, according to which cognition is based on the interaction of a multitude of processing units with changeable, weighted connections, specifically in terms of neural networks. It is therefore somewhat paradoxical that the new AI hype is mainly triggered by the success of neural network developments such as DL.[3] The new notion of AI that more and more establishes can be seen as an umbrella term covering all major approaches that aim to develop intelligence or cognition in machines or artificial systems. ML, in turn, is one of the new AI's major subfields with artificial neural networks as a further subfield. Among neural networks, DL networks represent the latest and most successful development. DL architectures comprise various types of deep neural networks such as feedforward, recurrent and convolutional neural networks (see the standard textbook by Goodfellow et al. 2016 for an overview as well as López-Rubio 2018 and Buckner 2018, 2019 as first systematic reflections on DL in the philosophy of science and mind literature).

It is instructive to divide the history of neural networks or connectionism into three historical phases or waves, each interrupted by a characteristic phase of stagnation:

- *1st wave: 1950 – early 1960s*. Early connectionism produces simple feedforward networks, especially the perceptron. Its developer, Frank Rosenblatt, is the decisive pioneer of this phase.

---

3  Compare a telling quote by Geoffrey Hinton (in: Ford 2018, p. 77-78): "For most of my career, there was artificial intelligence, which meant the logic-based idea of making intelligent systems by putting in rules that allowed them to process symbol strings. […] Then there was this other thing that wasn't AI at all, and that was neural networks. […] The logic-based people were interested in symbolic reasoning, whereas the neural network-based people were interested in learning, perception, and motor control. They're trying to solve different problems […]
   What's happened now is that industry and government use "AI" to mean deep learning, and so you get some really paradoxical things. In Toronto, we've received a lot of money from the industry and government for setting up the Vector Institute, which does basic research into deep learning, but also helps the industry do deep learning better and educates people in deep learning. Of course, other people would like some of this money, and another university claimed they had more people doing AI than in Toronto and produced citation figures as evidence. That's because they used classical AI. They used citations of conventional AI to say they should get some of this money for deep learning, and so this confusion in the meaning of AI is quite serious. It would be much better if we just didn't use the term 'AI.'"

- *„Neural winter"*: Minsky & Papert's (1969) criticism on perceptrons was so influential that it led to about 15 years of stagnation in the field of research and development of neuronal networks. During this time, the GOFAI paradigm was dominant. Its limitations, especially regarding pattern recognition as an important practical application, led at the end of the 1970s to the decline of this heyday of classical symbolistic AI and thus to the end of the neural winter.[4]

- *2nd wave: 1980 – early 1990s*. The 1980s represent a powerful return of (neo-) connectionism with a flood of novel neural network models. These include, in particular, Hopfield networks as auto-associative memories, Boltzmann networks, backpropagation, self-organizing feature maps (especially Kohonen networks) and recurrent networks. And even the spiking networks appearing in the early 1990s, although using a novel, time-dependent plasticity, belong to this phase.

- *"2nd neuronal winter": 1990 – early 2000s*. The renewed decline of connectionism in the 1990s can largely be attributed to a general problem that could well be called the „scaling problem": neuronal models designed for a manageable number of neurons (about 10-100) cannot so easily be scaled up to millions or billions of neurons without divergence problems. Another open question was whether wide nets (with high numbers of neurons per layer) or deep nets (increase in the number of layers) are to be preferred.

- *3rd wave: from 2010 onwards*. Three factors made today's third wave of connectionism possible under the buzzwords "machine learning", "deep learning" or quite simply "AI" (see also Brockman 2019, Ford 2018, Sejnowski 2018): firstly, various achievements at the end of the 2000s and beginning of the 2010s that made DL networks mathematically controllable and feasible (cf. Hinton & Salakhutdinov 2006, Krizhevsky et al. 2012), secondly the rapid development of computing power (such as the development of fast GPUs, i.e. graphics processors) and thirdly the huge amount of available training data only made possible by the Internet (this is an important point of contact between the developments of AI and Big Data).

---

[4] There are different notions of "winter" periods in the history of AI. In line with what was stated in the beginning of this section, the different schools and protagonists in the history judge different periods differently. The term "AI winter", for instance, is often used by GOFAI proponents to indicate the decline of research funding in the second half of the 1980s, while AI had been rather strong before. Almost the reverse is true for connectionism: while neural network research was down in the 1970s, the 1980s brought PDP and neo-connectionism back on stage. For the above usage of the term "neural winter" compare Sejnowski (2018, pp. 1,35) for the first neural winter and Bengio (in: Ford 2018, p. 25) for the second.

## 2.3 General AI

While AlphaGo still relied on human expert knowledge during its initial training phase, DeepMind was able to overcome this weakness just one year later with the development of AlphaGoZero. AlphaGoZero defeated AlphaGo in 2017 with an overwhelming 100:0. The accompanying Nature paper speaks of "Mastering the game of Go without human knowledge" (Silver et al. 2017), because AlphaGoZero is a self-learn system from scratch (modulo a prescription of the Go rules). For the first time ever, it was possible to create a machine that masters a highly complex cognitive task through self-learning alone. But the system does much more than that: AlphaGoZero masters Go on a super-human level that leaves the best human experts by orders of magnitude behind. According to its developers, AlphaGoZero had already reached the playing strength of the Lee Sedol version of AlphaGo after 4.9 million training games in three days of self-training.

In addition to the ability to self-learn (almost) without a priori knowledge, the second outstanding feature of AlphaGoZero is that it is able to acquire additional skills through self-learning without any change in the basic architecture. Unlike DeepBlue, which is an ANI system, AlphaGoZero's capabilities are potentially generalizable within a large domain of tasks. In December 2017, DeepMind introduced the further developed system AlphaZero (Silver et al. 2017). It masters Go, Chess and Shogi on a superhuman level after only a one-day training phase. Other recent developments in the Alpha series include AlphaFold, a system for predicting the 3D structure of proteins, and AlphaStar, a system mastering StarCraft II, one of the most challenging real-time strategy games in e-sports (see deepmind.com, August 2019 release). StarCraft II is even more complex than Go and it also demands dealing with incomplete information, and here again human top players have meanwhile been beaten. These are remarkable steps towards General AI (AGI: artificial general intelligence) – the level of machine intelligence that is equal to human intelligence with regard to any task. AGI is thus very often considered to be equivalent to human intelligence (HAI: human-level AI).

But of course AlphaZero is still a long way from real AGI. In particular, the specification of a special target function, with respect to which the system then allows generalizability, is still a clear limitation. Nevertheless, DeepMind's developments showed from the outset a remarkable potential for generating creative solutions and not just short-term success-oriented strategies. For example, the 37th move in the second game by AlphaGo against Lee Sedol, in which the system violated a millennium-old Go wisdom, was seen by the experts as spectacular. It not only confused the grandmaster, but also soon proved to be crucial for AlphaGo to win the game.

A creative solution that is easier to understand but no less original is already evident in one of the predecessor systems. DeepMind's first AI, a reinforcement learning based system called Deep Q-Network (DQN), successfully learned to play various classic Atari computer games (Mnih et al. 2013). The game "breakout", for example, is about hitting a virtual stone wall in a kind of 2D-squash-court-setting with racket and ball, and to remove a stone with every wall hit and score points, until finally the whole wall disappears. DQN independently learned the tricky strategy of drilling a tunnel through the wall and pushing the ball into the back of the wall, causing it to remove large amounts of stones (as DeepMind founder Demis Hassabis reports, the developers did not know this trick, which is well-known among Atari gamers; cf. Tegmark 2017, p. 122). Let us finally compare the most important characteristics of DeepBue and AlphaZero:

| System | Architecture | Strategy | Restriction |
| --- | --- | --- | --- |
| DeepBlue | GOFAI | rule-based | ANI |
| AlphaZero | DL network | self-learning | potentially generalizable |

## 3. New questions in the philosophy of AI

The recent developments in AI have not yet found the full attention from philosophy of science and philosophy of mind and cognition that they deserve (notable and most recent exceptions are Buckner 2018, 2019, López-Rubio 2018, Páez 2019, Schubbach 2019, and Zednik 2019). Some of the well-know foundational arguments and considerations in these fields of philosophy are gaining new relevance in view of the recent developments in the third wave of connectionism. Let us turn our attention to these arguments and considerations.

### 3.1 From the blockhead objection to the black box problem

Ned Block (1981) formulated an obvious objection to the Turing machine functionalism of classic AI that became prominent under the term "blockhead". The blockhead objection aims at the fact that an AI program can very well show intelligent behavior without being truly intelligent. Consider, for example, a regular conversation in a given language. It will always get along with a maybe high, but nevertheless finite number of sentences that stand in logical relationships to each other. This, in principle, allows for the possibility that a machine system, the blockhead program, is able to have a conversation by simply retrieving a pre-programmed look-up table of sentences.

Block's objection touches on two aspects. First, a critique of a purely behaviorist understanding of intelligence and cognition. Second, the question of what role the internal structure of an intelligent system plays. Both aspects are interrelated. According to Block, a test of cognitive abilities and intelligence based solely on external behavior, such as the Turing test (see section 3.4), is inadequate, as there are obviously internal structures that do not produce intelligence.

It is rather evident that the blockhead objection can be directly applied to a GOFAI system like DeepBlue, as the system merely uses a look-up tree algorithm and built-in heuristics based on human expert knowledge. Does DeepBlue really play chess? The question can be denied with good reasons already on the syntactic level: DeepBlue does nothing more than just rule-based manipulation of internal symbols. The question of whether DeepBlue has an understanding of chess in the sense of grasping the meaning of the chess pieces and the game as a whole does not even have to be considered. A system that only predicts combinatorial possibilities operates even syntactically not like a human chess player, who, due to lack of computational power, cannot heavily rely on combinatorics, but has to invoke self-learned and self-trained intuition and heuristics (whatever these psychologically vague terms exactly mean).

Applied to a self-learning, flexible system like AlphaGo, which is in part also capable of creativity, the blockhead objection is much less plausible. It is, in fact, inappropriate precisely to the extent that the system fulfills the condition of self-learning. DL systems in general are largely free of programmed specifications. But what internal structures, logics or heuristics do these systems use? At this point a novel problem arises that affects much of the developments of the new wave of connectionism. The internal structure of AlphaGo and related DL systems is only known in outline. This is a consequence of the deep structure and complexity of these systems as well as their ability to self-learn. This is known as the problem of *opacity* or the *black box problem* of deep neural networks.

The black box problem has led to a unique and novel sub-discipline in machine learning called *Explainable AI* (or XAI for short). It is driven by two motivations: on the one hand, the developers want to understand their own systems, on the other hand, the use of DL systems is in many areas – particularly obvious in the case of AI-based decision aids in medicine or self-driving cars – sensitive to whether and to what extent one is able to account for the question of how the systems do what they do. XAI therefore focuses, for example, on the development of tools of visualization and analysis that allow to open the black box and to understand the internal mechanisms of DL or related machine learning systems step-by-step and by means of reverse engineering. In computer science, a remarkable interest has emerged to explore the methodology in the field of machine learning and thus to practice a kind of philosophy of science.

## 3.2 The Symbol grounding problem

It is immediately obvious from the blockhead objection that DeepBlue doesn't really play chess. Essentially, the system won by calculating a huge look-up tree of possible moves with brute computational power (supplemented, of course, by effectively using a complex evaluation function and a Grandmaster game database). By no means, however, had the system the ability to learn. By way of contrast, for a rigorous self-learn system with black box character like AlphaGo (or AlphaGoZero) the blockhead objection cannot be decided so easily. The rest of this second part of the paper is devoted to the central question: *Does AlphaGo actually play Go?* This question will lead us to classical topics such as the Turing test, the symbol grounding problem, the Chinese room argument, and fundamental meaning-theoretical considerations.

Let us consider DQN, as already briefly described, and its ability to play Atari breakout on a super-human level. As typical for DL systems, DQN draws on a gigantic amount of training examples. In fact, the number of training games exceeds human training and thus the experience of human Atari gamers by orders of magnitude. This already suggests that the DL algorithms and network architectures do not strictly correspond to those that play a role in humans. Hence, DL systems used so far are biologically realistic in their connectionist surface structure only, but not in a strict sense. But this makes them no less successful with regard to certain capabilities, and they can nevertheless be regarded as a "proof of principle" for a biologically inspired connectionism.

Let's take a closer look at the training data of DQN. How can they be understood *from the perspective of the system*? For a human player, breakout's block-like pixel world, despite its minimalism, looks like a world of rackets, balls, walls and stones. There is nothing to suggest that this is the case for DQN. The system never had contact with real rackets, balls, walls or stones. From DQN's perspective, the only things that exist, as it were, are tons of pure pixel distributions. With that in mind, the system's superhuman playing abilities, and especially its additional ability to develop creative long-term solution strategies (such as digging a tunnel), become almost spooky. The same applies to AlphaGo or AlphaGoZero.

The problem can be seen as an instance of the symbol grounding problem, although this problem, in its original form, is aimed at classical symbolism. A symbol is a physical token that is individuated based on its physical form and that can be linked to other symbols according to syntactic rules. Symbols are therefore elements of symbol systems. Symbolism regards the manipulation of physical symbols as necessary and sufficient for intelligence and cognition. This is compatible with a computational theory of mind according to which the brain as the carrier of cognition is to be regarded as a computer or Turing machine. According to Stevan Harnad the symbol grounding problem consists in the following: „*Suppose you had to learn Chinese as a first language and the only source ... you had was a Chinese/Chinese dictionary! … How is symbol meaning to be grounded in something other than just more meaningless symbols? This is the symbol grounding problem*" (Harnad 1990, p. 339-340).

In contrast to symbolism, connectionism emphasizes not only the network architecture of cognitive systems, but also a "subsymbolism" instead of symbol-based information processing. Superficially, neural networks do not operate on symbols such as Chinese characters or zeros and ones, but on inputs that represent features. In the case of DQN or AlphaGo, these are pixels with different gray or color values. However, since it can be shown that important classes of neural networks such as recurrent networks are Turing complete (cf. Siegelmann & Sontag 1995), these systems ultimately also operate symbolically insofar as they can be mapped to the symbolic operations of Turing machines. The question of whether and how the input pixel distributions are meaningful for DQN or AlphaGo amounts to the question as to what extent these distributions have a "grounding" or "anchoring" in the world. And superficially, it seems as if one has to say that they do not have any such grounding. Therefore, neither DQN nor AlphaGo operate meaningfully, they do not understand what they are doing.

But this conclusion falls short because it overlooks an important distinction regarding meaning. Consider chess. How do the chess pieces get their meaning? What makes a knight a knight? Two things: It must be different in shape from all other types of pieces, such as rook or bishop. It then acquires its meaning in the game through exactly the role it plays in the game, which in turn is clearly assigned to it by the rules of the game. It is therefore a physical symbol that is manipulated according to rules – and the semantics of this symbol originates from the syntax of the game. In contrast, consider the words of a spoken language. They allow for sequences according to grammatical rules to form sentences. However, the question of what a particular word, such as the word "tree," refers to, is in no way determined by the grammar. Semantics in the sense of reference does not originate from syntax.

We must therefore distinguish between meaning in the sense of *functional role*, which is determined by internal rules, and meaning in the sense of *external reference*. The symbol grounding problem primarily asks for meaning in the second sense: how can system-internal symbols be grounded in the world so that they acquire a meaning in the sense of reference?

Consider again the example of Atari breakout. DQN seems not to dispose of the meaning of the terms racket, ball, etc. in the sense of reference. However, it is by no means excluded that the learning performance of DQN consists essentially in the fact that it recognizes certain stable and recurring patterns in pixel distributions and links them to regular behavior. A sufficient XAI analysis could provide exactly this kind of information, as it would have to be shown that DQN represents stable pixel configurations in higher layers and thus achieves the concepts racket, ball etc. in the sense of a *functional role semantics* (FRS).

One might reasonably assume that, for the purposes of significance *in* the game, everything essential has been achieved by an FRS framework. For if we look for instance at AlphaGo, the question of semantics in the sense of reference does not arise at all, since Go pieces just like chess pieces have no reference to things or states of affairs in the world, but only an internal functional role within the system, i.e. a meaning *in* the game.

## 3.3  The Chinese room argument

Harnad's symbol grounding problem was inspired by Searle's related and well-known Chinese room argument (Searle 1980, 1990, cf. Harnad 1989, 2001). In a way, Harnad's argument makes the deeper core of the Chinese room argument explicit. The latter argument aims to show that syntax is not sufficient for semantics, that the human brain is not a computer in the sense of symbolism, and that the computational theory of mind is therefore wrong. To this end, Searle conceives the Chinese room as the caricature of a Turing machine, where he himself takes over the role of a tape head for reading and writing by sitting in an otherwise empty room and by using a set of rules (the machine table) provided to him and allowing him to manipulate Chinese characters that he obtains in the room as input and that he reaches out as output. Since Searle doesn't understand Chinese and since for him Chinese symbols look like "meaningless squiggles", he insists that he can never attain the meaning of Chinese symbols in this way, i.e. by pure syntactic symbol manipulation.

In the 1980s, the Chinese room argument triggered a flood of reactions and discussions, among the most common objections to the argument are the connectionist critique (Searle 1990) and the criticisms referred to by Searle as *systems reply* and *robot reply* (Searle 1980). Searle basically counters all objections according to the same strategy by showing that they just provide "more of the same". According to the systems reply the whole room rather than the internal operator is proficient in Chinese. Searle, however, argues that he could just as well internalize the whole room (particularly by learning the rule book) and still do nothing but mere syntactic symbol manipulation. According to the connectionist variant of the systems reply we are asked to consider an entire network of operators rather than a single operator. Searle, again, argues that we can as well imagine a "Chinese gym" with lots of operators manipulating symbols according to rules, but that still neither any of the operators nor the whole gym would thereby acquire the meaning of Chinese symbols.

Of particular importance is the robot reply. Would not a robot equipped with the rules of Chinese operating in the Beijing marketplace gain the meaning of the previously merely syntactic symbols in the course of this interaction? Wouldn't a system in this way establish the necessary reference-to-world relation? According to Searle, this is not the case, since the "computer inside the robot" (Searle 1980, 420) is still an analogue of the Chinese room. Be it that the input stems from an external camera and the output is used to control the arms, on the level of the internal computer that controls the robot both input and output still consist of nothing but mere meaningless symbols. This answer is reminiscent of a strange homunculus conception, and the question also arises as to whether a combination of systems and robot reply cannot be reinforced by further arguments from the areas of embodied and situated cognition (cf. Robbins & Aydede 2009).

But we do not need to pursue this further here. Since Searle's argument is an argument against GOFAI's symbolism (for Searle: "strong AI"), it merely aims to show that meaningful thinking goes beyond algorithmic symbol manipulation. But if, in order to attain semantics, embodiment, social interaction and situatedness are crucial, this even ultimately strengthens the argument. And it shows: in essence, the discussion of the Chinese room argument boils down to the symbol grounding problem: how can meaning in the sense of reference be grounded?

Strangely, the technologies now available in the area of machine language translation through deep learning seem to realize the Chinese room scenario, at least in part. Freely available systems such as Google Translate or DeepL have shown a breathtaking improvement in their translation performance in recent years. And yet: one would hardly want to assume that one of these systems truly understands the texts they translate, sometimes in excellent quality. The new systems go beyond earlier forms of either rule-based or statistic-based machine translation. They extract rules of word selection, word order etc. by self-learning on the basis of voluminous bilingual text corpora.[5] All this suggests the following: syntax is 'almost sufficient' to produce the linguistic behavior that corresponds to the behavior of speakers who truly possess semantics. Although syntax is not completely sufficient for semantics, syntax is 'almost sufficient' in the sense that it is sufficient for all practical purposes (but still insufficient from a strict Searlean point of view). This means that, in effect, a syntactic machine can be indistinguishable in its translation performance from a human speaker.

---

5   A Google representative told George Dyson: "We are not scanning all those books to be read by people... We are scanning them to be read by an AI" (interview in Brockman 2019, p. 64).

In the light of the above considerations, it follows that DL translation systems do not appear to have any relation to the world, their internal symbol processing is "ungrounded". Nevertheless, these systems acquire rule knowledge and meaning in an FRS sense through self-learning. This alone is remarkable. The crucial step, however, is yet to come. How can it even be possible that a mere symbol based FRS becomes effectively indistinguishable (i.e. regarding language behavior) from a truly referential semantics? Should it be possible that, when it comes to semantics, we can do with a pure FRS alone and dismiss any appeal to reference? On the face of it, this amazing possibility seems to be suggested by the translation capabilities of systems such as Google Translate and DeepL. But on closer inspection, it's not. The systems do indeed go beyond a pure FRS. The text corpora used in learning create a kind of indirect world reference. They were created by human speakers who dispose of a semantics in a referential sense. Hence, Google Translate and DeepL have no direct but an *indirect grounding*, they refer indirectly to the world. This is probably no sufficient form of reference in the sense of a genuine causal theory of reference (following Fodor 1987; see also section 3.5), but such a theory may be too strong in its requirements anyway. Regularities that can be extracted from text corpora comparisons go beyond mere grammatical regularities, they also provide world regularities insofar as the texts deal about the world. This allows to extract a decent amount of structural information about the world.[6]

## 3.4 The Turing test

Alan Turing's proposal to use an imitation game to operationalize intelligence has always been controversial. According to Turing (1950), if a machine succeeds in being indistinguishable from a human being in its response to arbitrary questions, then intelligence and higher cognitive abilities should be attributed to it. Searle's Chinese room cannot only be understood as an objection to symbolism, but also as an objection to the Turing test (conversely, this does not mean that symbolists are committed to the Turing test – symbolism as a doctrine of cognition and the Turing test as an external test procedure for intelligence are independent).

---

6   It may not allow to extract information about the intrinsicality of things in a rigorous ontological sense. According to the doctrine of structural realism, however, such a concept of intrinsicality is in conflict with our best knowledge about the bottom level and can therefore be regarded as doubtful anyway; cf. Lyre 2010.

Numerous examples of Turing-like scenarios indicate the weakness and limitations of this test procedure. Weizenbaum's (1976) early experiences with his well-known imitation program ELIZA, which was able to conduct a psychotherapeutic dialogue, are telling (and were frightening for Weizenbaum): ELIZA was based on comparatively simple scripts and structured dictionaries, yet some test persons could not escape the impression of a conversation with a real psychotherapist. A more recent example is the Goostman chatbot. It scored a surprisingly good pass on several Turing test contests in the early 2000s. The bot simulates a 13-year-old Ukrainian boy, taking advantage of the fact that people more easily concede grammatical mistakes and lack of general knowledge to such a personality. These examples show that the Turing test offers no sufficient criterion for meaningful cognition, since it can be passed with too simple and possibly also "dishonest" means (as, for instance, in the form of a blockhead).

Assistance systems such as Siri, Cortana or Google Assist provide the contemporary variant of Turing-like scenarios. At its annual developer conference I/O 2018, Google surprised the general public with presenting Google Duplex, a system currently under development. It is meant to support everyday life, for example by making appointments for the user. Google had tested its system in real life by scheduling a restaurant reservation or calling a hairdresser to book an appointment. The natural-language performance of the system is shockingly good: the called persons could not have guessed that they actually spoke to a machine. The phone calls were fluent and spontaneous including prosodic and non-verbal elements such as "hmm" and "uh" together with natural intonation and breaks. This provides another example of successfully passing the Turing test, this time of a DL-based AI.

Of course, the ultimate goal is to generalize the Turing test along the axis of the generalization capabilities of an AI system, i.e. in view of the system's scope in problem-solving. A Turing test for AGI would also require the system to have practical skills such as active navigation through natural environments, producing or repairing things, or social interaction and activities (see also the final section). Nevertheless, any Turing test remains on the level of purely external functionality, an insight into the internal goings-on of the black box is against the behaviorist spirit of the test. This, however, is in strong contrast to the intuition of the blockhead objection according to which an understanding of the internal structure and mechanisms of an AI system is indispensable to assign cognitive or intelligent properties to it. Therefore, the Turing test should not be regarded as a sufficient criterion for intelligence. It may still be regarded as a necessary criterion: AI systems should perform functionally and behaviorally equivalent to humans in order to be regarded as cognitive. Sufficient for this attribution, however, is an understanding of the relevant internal structure of the system, at least to a certain extent. To which extent can presumably not be generally said in advance, but it must at least be possible to open the black box in part. And by this, it must then also be possible to provide information about the systems grounding, i.e. whether and how it refers to the world.

### 3.5 Meaning as use and social practice

Our considerations have already led to tricky questions about the concepts of reference and grounding. How far should one go with these questions? This depends on how far one wants to build up one's theory of meaning primarily as a referential semantics. The standard approach in this domain are causal theories of reference (cf. Fodor 1987). However, such theories lead to known difficulties in connection with questions of misrepresentation. In short: if one ties the meaning or semantic content of a mental representation strictly to its cause, then by definition there is no leeway for misrepresentation. This, however, sounds disquieting.

A common approach to non-referential semantics is provided by use of theories of meaning, as first outlined by Ludwig Wittgenstein (1953). The central idea is to trace meaning back to linguistic use and social practice. According to Wittgenstein, the diversity of language can be seen in the variety of ways in which it is used. The focus is on the concept of rules. A classical, rule-based conception of language sees language as regulated by some unambiguous syntax. This applies all the more to formal languages or mathematics (and has tacitly been assumed in our previous considerations on the relationship between syntax and semantics). As Wittgenstein aims to show in his "rule following" considerations, such a strict Platonic conception of rules leads to an infinite regress. In order to set up the syntactic rules of, say, a certain Turing machine, other rules are required governing the former rules. But these too satisfy further rules, hence a regress follows.

According to Wittgenstein, language is limited by rules, but these rules presuppose a public practice and only become apparent in use. The rule following problem consists in the fact that language use and practice are always finite, but that no finite number of cases determines the "rules" of language use and thus the meaning of linguistic expressions under *all*, hence infinitely many, circumstances. Language rules are by no means rigid, but depend on the social context. Wittgenstein's bizarre thought experiment of the two-minute-man drastically demonstrates the consequences of his conception:

> *„Let us imagine a god creating a country instantaneously in the middle of the wilderness, which exists for two minutes and is an exact reproduction of a part of England, with everything that is going on there in two minutes. Just like those in England, the people are pursuing a variety of occupations. Children are in school. Some people are doing mathematics. Now let us contemplate the activity of some human beings during these two minutes. One of these people is doing exactly what a mathematician in England is doing, who is just doing a calculation. - Ought we to say that this two-minute-man is calculating? Could we for example not imagine a past and a continuation of these two minutes, which would make us call the process something quite different?"* (Wittgenstein 1956, VI §34)

Wittgenstein's answer is obvious: the two-minute human "does not calculate" because he is not embedded in the practice and context of mathematics. Against this backdrop, let us consider our guiding question whether AlphaGo actually plays Go. Games, like language, are limited by rules. Wittgenstein insinuates a tight analogy between games and language and between the corresponding roles of rules and rule use. Indeed, he speaks of language as a "language game". Just as there is no mathematics or linguistic meaning without a social context, there are no games. Hence, from a Wittgensteinian understanding of use and practice, AlphaGo does not play Go since it lacks social context: the shared and public practice of the game of Go.

In section 3.2 our conclusion was that the functional roles comprise everything that's essential in terms of meanings *in* the game. The main reason for this was that the meaning of moves and pieces is not referential, as for instance chess pieces do not refer to anything in the world. Following Wittgenstein, however, the meaning of games still has a kind of grounding, even if not in the referential sense. Instead, it is a kind of social grounding. Without public practice rules of games won't be subject to external control and, therefore, are no rules at all.

Wittgenstein's reflections on rule following are undoubtedly radical (and accordingly controversial), as the two-minute-man scenario drastically demonstrates. Saul Kripke saw himself prompted to a radical rule skepticism, which infects not only rules of games or language, but even the rules of mathematics and logic (Kripke 1974). An in-depth discussion of these questions is far beyond the scope of the current paper. We shall assume that, for all practical purposes, rule knowledge can be set up in an AI machine modulo "Kripkensteinian" doubts.

Thus, for each version of the Alpha series, from AlphaGo to AlphaZero, the respective rules of the games to be learned were unambiguously be implemented (Silver et al. 2017). The machine then develops a functional role semantics about the elements and overall setup of the game limited by these pre-determined rules. The systems of the Alpha series have no further grounding. Google Translate or DeepL, on the other hand, already have a rudimentary form of a socially anchored semantics, because these systems acquire an indirect social grounding in the course of their translation learning. After all, the text corpora on the basis of which the systems learn were generated by socially situated speakers, and are therefore parasitic with regard to their social practices. A future AI that combines, for example, the external performance of Google Duplex with the indirect grounding of world knowledge on the basis of Internet data could ultimately become a real part of our social practice of language and, hence, a real part of the language community. There is no convincing reason to assume that such systems would still lack a proper semantic grounding.

## 4. The state space of AI

The extensive discussion of the two preceding sections should now help us to achieve the overall objective of the paper: to develop a general model of the state space of AI. Our main question is: what are the dimensions that span this space? In section 2, the main features of the "new AI", the third wave of connectionism, were presented. In section 3, newly posed questions of the philosophy of AI were discussed. The central motif of the section 2 was the ability of self-learning, the motif of section 3 was grounding, i.e. the semantic anchoring of AI systems in the world. This suggests a first answer to our question: *self-learning* and *grounding* represent two essential dimensions of the AI state space. In section 2, we also came across a third dimension: the notion of *generalization* takes into account whether an AI system is limited to a specific problem or problem class, or whether it can be extended and generalized to some wider domain of tasks. Accordingly, the AI state space is to be conceived as a three-dimensional space spanned by the dimensions:

   - Generalization (from narrow to general AI),

   - Self-learning (from rule-based to learn-based),

   - Grounding (degree of semantic world reference).

Let us locate some of the mentioned AI systems in this space (see figure 1). A classic GOFAI system like DeepBlue is close to the origin. GOFAI systems are rule-based rather than learn-based, and almost all of them are narrow AI systems (DeepBlue, for instance, is confined to chess as a special problem). Moreover, typical GOFAI systems dispose at best about an internal FRS (as DeepBlue does in terms of the functional roles of chess pieces). AlphaGo, on the other hand, sits at a much higher position in the self-learning dimension, we then arrive at AlphaGoZero and AlphaZero by successive shifts parallel to the generalization axis. None of the mentioned systems, however, has a semantic grounding beyond an internal FRS. At best, AI assistance systems such as Google Duplex move into this dimension, albeit still weakly at present.

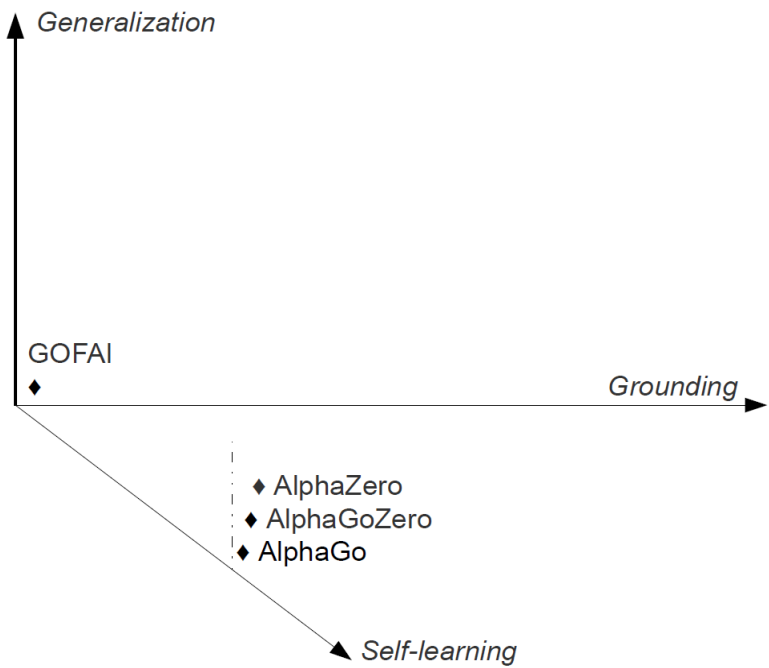

[Figure 1]

It would be desirable to proceed from the state space topology (dimensionality and neighborhood) to a metric space (to determine distances). Human-level AI (HAI) is a point of orientation. HAI has values in all dimensions and can therefore be used to calibrate the coordinate axes. In addition, systems that lie on the extended radial connecting line between origin and HAI (or within a suitably chosen spatial angle range) mark the area of superintelligence or superhuman AI (SAI) (see figure 2).

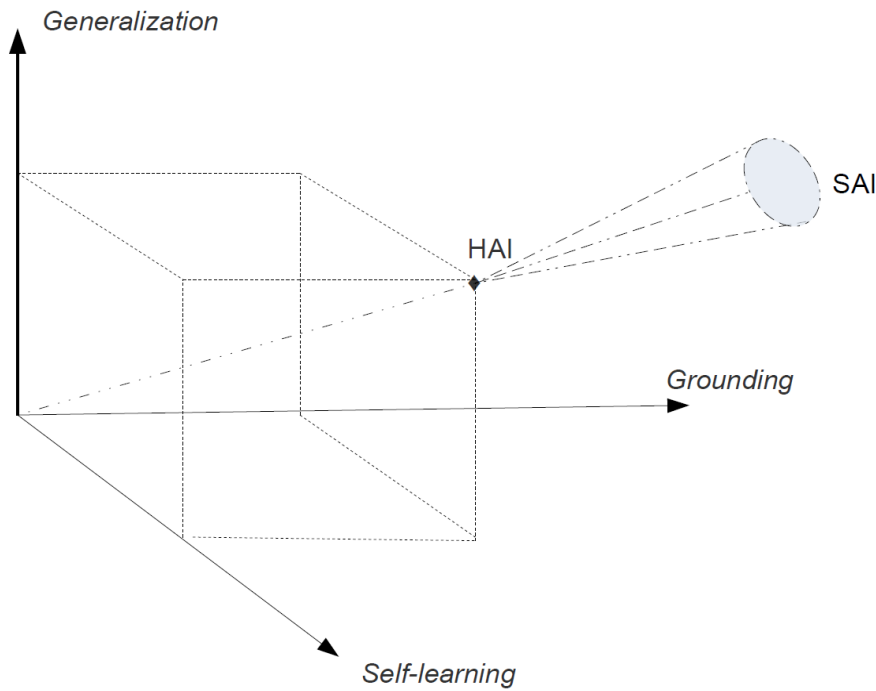

[Figure 2]

A detailed determination of the metric goes beyond the scope of this paper and is the task of further investigations. Let us, instead, focus once again on the dimensionality. While the three-dimensionality of the AI state space offers a first orientation, it is strictly speaking an approximation in terms of a simplified dimensionality reduction. The point is that, on closer inspection, two of the three dimensions decompose into further subdimensions. They could be dubbed "main dimensions", as they actually represent subspaces of the AI state space. While the generalization dimension is already correctly identified, grounding is one such main dimension that we will discuss in more detail below.

The dimension of self-learning needs some unpacking. Self-learning determines as to what extent an AI system can change its own configuration by means of self-organization. There are, however, further dimensions of self-organization that are crucial for the development of intelligence and cognition. The comparison with biological intelligent systems suggests self-repair and self-replication as additional dimensions. In general, we should exploit the so-called *self-x qualities* in the spirit of the program of organic computing. The key idea of organic computing is that self-organization or self-organized system configurations play a decisive role for AI (cf. Würtz 2008). Self-learning is only one of the self-x properties. Depending on the author, the self-x properties in organic computing include properties like self-learning, self-configuring, self-optimizing, self-healing or self-protecting. Some of these properties are not quite distinct (e.g. self-healing and self-protecting), others are vaguely subsumed under self-configuring, although strictly speaking all self-x properties are properties of the configuration (from this perspective, self-configuring and self-organizing are, as it were, "meta self-x properties"). Let us, as a heuristic suggestion, consider the qualities

- self-learning,
- self-repairing,
- self-replicating.

Self-x properties in general represent the main dimension we are looking for, i.e. they span the self-x subspace of the AI state space. An exhaustive identification of all relevant self-x properties and a necessary analysis of the terms configuration and self-organization goes beyond the scope of the current paper, but a somewhat anticipatory consideration should nevertheless be added here. Our concern is not just about organic self-organization, but about the most general self-x properties of any intelligent systems, whether artificial or biological. In this sense, the concept of AI just includes natural and especially human intelligence as special cases. It is therefore much more likely that further self-x properties must be accounted for. Recall section 3.4, where it was argued that a Turing test on AGI should include not only pure responsive but also practical skills, where the system is asked to actively explore its environment. Let us therefore add the corresponding self-x property to the above list:

- self-exploratory.

But even for Turing tests on AGI the black box problem essentially remains. In contrast, a qualitatively new level would be reached if AI systems were able to provide explanations or justifications of their own responses and actions. The ability of being self-explanatory would add to the above as a further significant self-x feature. And this immediately raises the question of whether AI systems should not also be self-understanding or even self-conscious. The list of self-x-properties may therefore be extended by the properties

- self-explanatory,
- self-understanding,
- self-conscious.

It becomes clear that a detailed analysis of the subspace of *self-x-ness* leads to far-reaching questions that go considerably beyond the scope of the current paper. For our purposes and as a main dimension, however, the self-x dimension of the AI state space is sufficiently characterized.

Let us finally turn to the main dimension of grounding, and thus to the corresponding AI subspace. Our discussion in section 3 has shown that grounding, understood in the general sense of bearing semantics, can in principle occur in at least three ways. First, a distinction must be made between an internal FRS and external semantics. Only the latter establish a genuine reference relation to the world. As two important candidates we considered referential semantics and use theories of meaning. Insofar as the semantic properties of AI systems denoted by these three ways are to be regarded as independent, the grounding dimension would be characterized by three subdimensions: functional role grounding, referential grounding, and social grounding.

Also, as any of the other dimensions, the grounding dimensions are understood as continuous dimensions. This is a further important point that can only be touched upon here. Semantic grounding isn't on-off. Intelligent systems, whether biological or artificial, may be more or less grounded. The semantic skills of apes outweigh the skills of ravens, which in turn outweigh the skills of ants. The semantic skills pertain the way in which intelligent beings are grounded or anchored in the world in terms of their meaningful grasp and understanding of that world. Humans, in turn, trump the semantic skills of any known animal. But the gradual differences in terms of grounding exist of course also within a species. Healthy human adults overtop the semantic skills of newborns or patients with dementia. Moreover, semantic grounding is open-ended. Future AI systems may likewise outweigh the semantic skills of humans. The consequences of this are largely unknown and speculative. This is one of the pressing questions and, presumably, big worries with the issues of singularity and superintelligence (Bostrom 2013, Tegmark 2017).

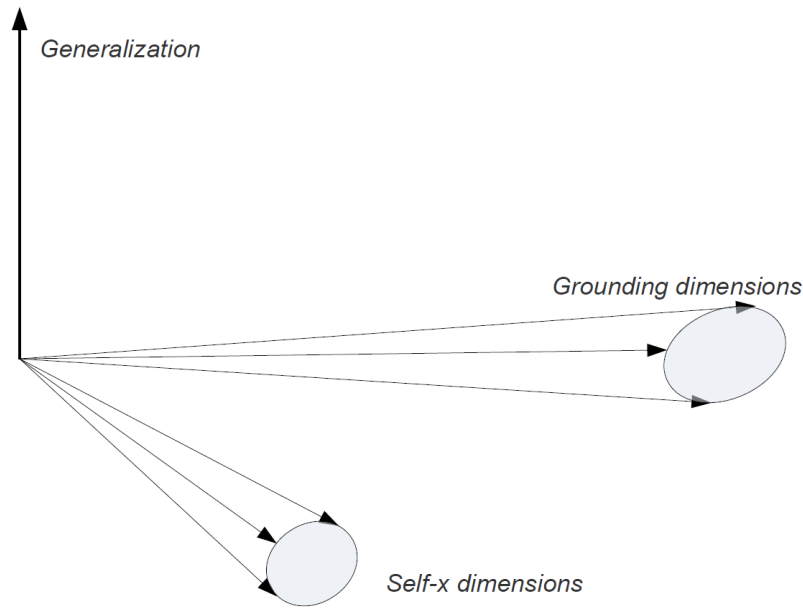

[Figure 3]

The thus developed model of the state space of AI (see figure 3) is certainly only a first draft. It remains to be examined whether further dimensions should be added or whether postulated dimensions should be changed or deleted. The usefulness of a general model of the state space of AI, however, is obvious. AI developments can be better classified and related to one another by locating them in a state space. The future development could also be taken more clearly into account. Indeed, not only the occupied regions of the state space are of importance, but also those sectors that can possibly never be reached by any AI are of fundamental interest. For example, it could very well turn out that there are no AI systems that have a high degree of generalization but are at the same time not self-learning or grounded to a certain extent. These and many other considerations are the task of further research and exploration of the proposed state space of AI.

## References

Block, Ned (1981): Psychologism and Behaviorism. *Philosophical Review* 90(1): 5-43.

Bostrom, Nick (2013): *Superintelligence. Paths, Dangers, Strategies*. Oxford University Press.

Buckner, Cameron (2018): Empiricism without Magic: Transformational Abstraction in Deep Convolutional Neural Networks. *Synthese* 195: 5339–5372.

Buckner, Cameron (2019): Deep learning: A philosophical introduction. *Philosophy Compass* 2019;e12625.

Brockman, John, editor (2019): *Possible Minds. 25 Ways of Looking at AI*. Penguin Press.

Chomsky, Noam (1980): Rules and Representations. *Behavioral and Brain Sciences* 3 (127): 1-61.

Fodor, Jerry (1987): *Psychosemantics*. MIT Press.

Ford, Martin, editor (2018): *Architects of Intelligence: The truth about AI from the people building it.* Packt Publishing.

Goodfellow, Ian, Yoshua Bengio & Aaron Courville (2016): *Deep Learning*. MIT Press.

Harnad, Stevan (1989): Minds, Machines and Searle. *Journal of Theoretical and Experimental Artificial Intelligence* 1: 5-25.

Harnad, Stevan (1990): The Symbol Grounding Problem. *Physica* D 42: 335-346.

Harnad, Stevan (2001): What's Wrong and Right About Searle's Chinese Room Argument? In M. Bishop & J. Preston (eds.): *Essays on Searle's Chinese Room Argument*. Oxford University Press.

Hinton, Geoffrey E. & Ruslan R. Salakhutdinov (2006): Reducing the dimensionality of data with neural networks. *Science* 313: 504-507.

Kripke, Saul A. (1982): *Wittgenstein on Rules and Private Language.* Harvard University Press.

Krizhevsky, Alex, Ilya Sutskever & Geoffrey E. Hinton (2012): ImageNet classification with deep convolutional neural networks. *Advances in Neural Information Processing Systems 25 (NIPS 2012)*, Vol. 1: 1097-1105.

LeCun, Yann, Yoshua Bengio & Geoffrey E. Hinton (2015): Deep learning. *Nature* 521: 436-444.

López-Rubio, Ezequiel (2018): Computational functionalism for the deep learning era. *Minds and Machines* 28: 667–688.

Lyre, Holger (2010): Humean Perspectives on Structural Realism. In: F. Stadler (ed.): *The Present Situation in the Philosophy of Science*. Springer, p. 381-397.

Mnih, Volodymyr, Koray Kavukcuoglu, David Silver, Alex Graves, Ioannis Antonoglou, Daan Wierstra & Martin Riedmiller (2013): Playing Atari with Deep Reinforcement Learning. Eprint *arXiv:1312.5602.*

Páez, Andrés (2019): The Pragmatic Turn in Explainable Artificial Intelligence (XAI). *Minds and Machines*, online first.

Robbins, Philip & Murat Aydede, editors (2009): *The Cambridge Handbook of Situated Cognition*. Cambridge University Press.

Schmidhuber, Jürgen (2015a): Deep Learning in Neural Networks: An Overview. *Neural Networks* 61: 85-117.

Schmidhuber, Jürgen (2015b): Deep Learning. *Scholarpedia* 10(11): 32832.

Schubbach, Arno (2019): Judging Machines. Philosophical Aspects of Deep Learning. *Synthese*, online first.

Searle, John R. (1980). Minds, brains and programs. *The Behavioral and Brain Sciences* 3: 417- 457.

Searle, John R. (1990). Is the Brain's Mind a Computer Program? *Scientific American* 1: 26-31.

Sejnowski, Terrence J. (2018): *The Deep Learning Revolution*. MIT Press.

Siegelmann, H. T. & E. D. Sontag (1995): On the computational power of neural nets. *Journal of Computer and System Sciences* 50(1): 132–150.

Silver, David et al. (2016): Mastering the game of Go with deep neural networks and tree search. *Nature* 529: 484-489.

Silver, David et al. (2017): Mastering the game of Go without human knowledge. *Nature* 550: 354-359.

Sutton, Richard & Andrew Barto (2018): *Reinforcement Learning: An Introduction*. 2nd edition. MIT Press.

Tegmark, Max (2017): *Life 3.0: Being Human in the Age of Artificial Intelligence*. Allen Lane.

Turing, Alan (1950): Computing Machinery and Intelligence. *Mind* 49: 433-460.

Weizenbaum, Joseph (1976): *Computer Power and Human Reason. From Judgement to Calculation*. W. H. Freeman.

Wittgenstein, Ludwig (1953): *Philosophical investigations*. Macmillan Publishing Company.

Wittgenstein, Ludwig (1956): *Remarks on the foundations of mathematics*. Blackwell.

Würtz, Rolf P., editor (2008): *Organic Computing (Understanding Complex Systems)*. Springer.

Zednik, Carlos (2019): Solving the Black Box Problem: A Normative Framework for Explainable Artificial Intelligence. Eprint *arXiv:1903.04361.*